\pdfoutput=1
\documentclass[11pt]{article}

\usepackage{naacl2021}

\usepackage{times}
\usepackage{latexsym}

\usepackage[T1]{fontenc}
\usepackage[utf8]{inputenc}

\usepackage{microtype}

\usepackage{amsmath}
\usepackage{amssymb}
\usepackage{graphicx}
\usepackage{multirow}
\usepackage{multicol}

\usepackage{amsthm}
\usepackage{booktabs}
\usepackage{algorithm}
\usepackage{algorithmic}
\usepackage{caption}
\usepackage{subcaption}
\usepackage{url}

\title{UniDrop: A Simple yet Effective Technique to Improve Transformer without Extra Cost}

\author{
	Zhen Wu\textsuperscript{\normalfont 1}\thanks{~~This work was done when Zhen Wu was a research intern at Microsoft Research Asia.} \quad
	Lijun Wu\textsuperscript{\normalfont 2} \quad
	Qi Meng\textsuperscript{\normalfont 2} \quad
	Yingce Xia\textsuperscript{\normalfont 2} \quad
	Shufang Xie\textsuperscript{\normalfont 2} \\
	{\bf Tao Qin}\textsuperscript{\normalfont 2} \quad
	{\bf Xinyu Dai}\textsuperscript{\normalfont 1} \quad
	{\bf Tie-Yan Liu}\textsuperscript{\normalfont 2} \quad
	\\ 
	\textsuperscript{1}National Key Laboratory for Novel Software Technology, Nanjing University \\
	\textsuperscript{2}Microsoft Research Asia \\
	\texttt{wuz@smail.nju.edu.cn, daixinyu@nju.edu.cn} \\
	\texttt{\{Lijun.Wu,meq,yingce.xia,shufxi,taoqin,tyliu\}@microsoft.com} \\
}

\begin{document}
\maketitle
\begin{abstract}
Transformer architecture achieves great success in abundant natural language processing tasks.
The over-parameterization of the Transformer model has motivated plenty of works to alleviate its overfitting for superior performances. With some explorations, we find simple techniques such as dropout, can greatly boost model performance with a careful design. Therefore, in this paper, we integrate different dropout techniques into the training of Transformer models. Specifically, we propose an approach named $\mathtt{UniDrop}$ to unite three different dropout techniques from fine-grain to coarse-grain, i.e., feature dropout, structure dropout, and data dropout. Theoretically, we demonstrate that these three dropouts play different roles from regularization perspectives. Empirically, we conduct experiments on both neural machine translation and text classification benchmark datasets. Extensive results indicate that Transformer with $\mathtt{UniDrop}$ can achieve around $1.5$ BLEU improvement on IWSLT14 translation tasks, and better accuracy for the classification even using strong pre-trained RoBERTa as backbone.
\end{abstract}

\section{Introduction}
In recent years, Transformer~\cite{vaswani2017attention} has been the dominant structure in natural language processing (NLP), such as neural machine translation~\cite{vaswani2017attention}, language modeling~\cite{dai2019transformer} and text classification~\cite{devlin2019bert,liu2019roberta}. 
To further improve the model performance, there has been much effort in designing better architectures or introducing external knowledge into Transformer models~\cite{wu2018pay,lu2019understanding,Kitaev2020Reformer,ahmed2017weighted,hashemi2020guided}, which increases computational costs or requires extra resources.

Despite the effectiveness of above strategies, the over-parameterization and overfitting is still a crucial problem for Transformer. Regularization methods such as weight decay \cite{krogh1992simple}, data augmentation \cite{sennrich2016improving}, dropout \cite{srivastava2014dropout}, parameter sharing \cite{dehghani2018universal,xia2019tied} are all widely adopted to address overfitting.
Among these regularization approaches, dropout \cite{srivastava2014dropout}, which randomly drops out some hidden units during training, is the most popular one and various dropout techniques have been proposed for Transformer. For example, \citet{fan2019reducing} propose $\mathtt{LayerDrop}$, a random structured dropout, to drop certain layers of Transformer during training. \citet{zhou2020scheduled} alternatively propose $\mathtt{DropHead}$ as a structured dropout method for regularizing the multi-head attention mechanism. Both of them achieved promising performances. One great advantage of dropout is that it is free of additional computational costs and resource requirements. Hence we ask one question: \textit{can we achieve stronger or even state-of-the-art (SOTA) results only relying on various dropout techniques instead of extra model architecture design or knowledge enhancement?}

To this end, in this paper, we propose $\mathtt{UniDrop}$ to integrate three different-level dropout techniques from fine-grain to coarse-grain, \emph{feature dropout}, \emph{structure dropout}, and \emph{data dropout}, into Transformer models. Feature dropout is the conventional dropout \cite{srivastava2014dropout} that we introduced before, which is widely applied on hidden representations of networks. Structure dropout is a coarse-grained control and aims to randomly drop some entire substructures or components from the whole model. In this work, we adopt the aforementioned $\mathtt{LayerDrop}$ \cite{fan2019reducing} as our structure dropout. Different from the previous two dropout methods, data dropout \cite{iyyer2015deep} is performed on the input data level, which serves as a data augmentation method by randomly dropping out some tokens in an input sequence.

\begin{figure*}
     \centering
     \begin{subfigure}[b]{0.35\linewidth}
         \centering
         \includegraphics[width=1.0\textwidth]{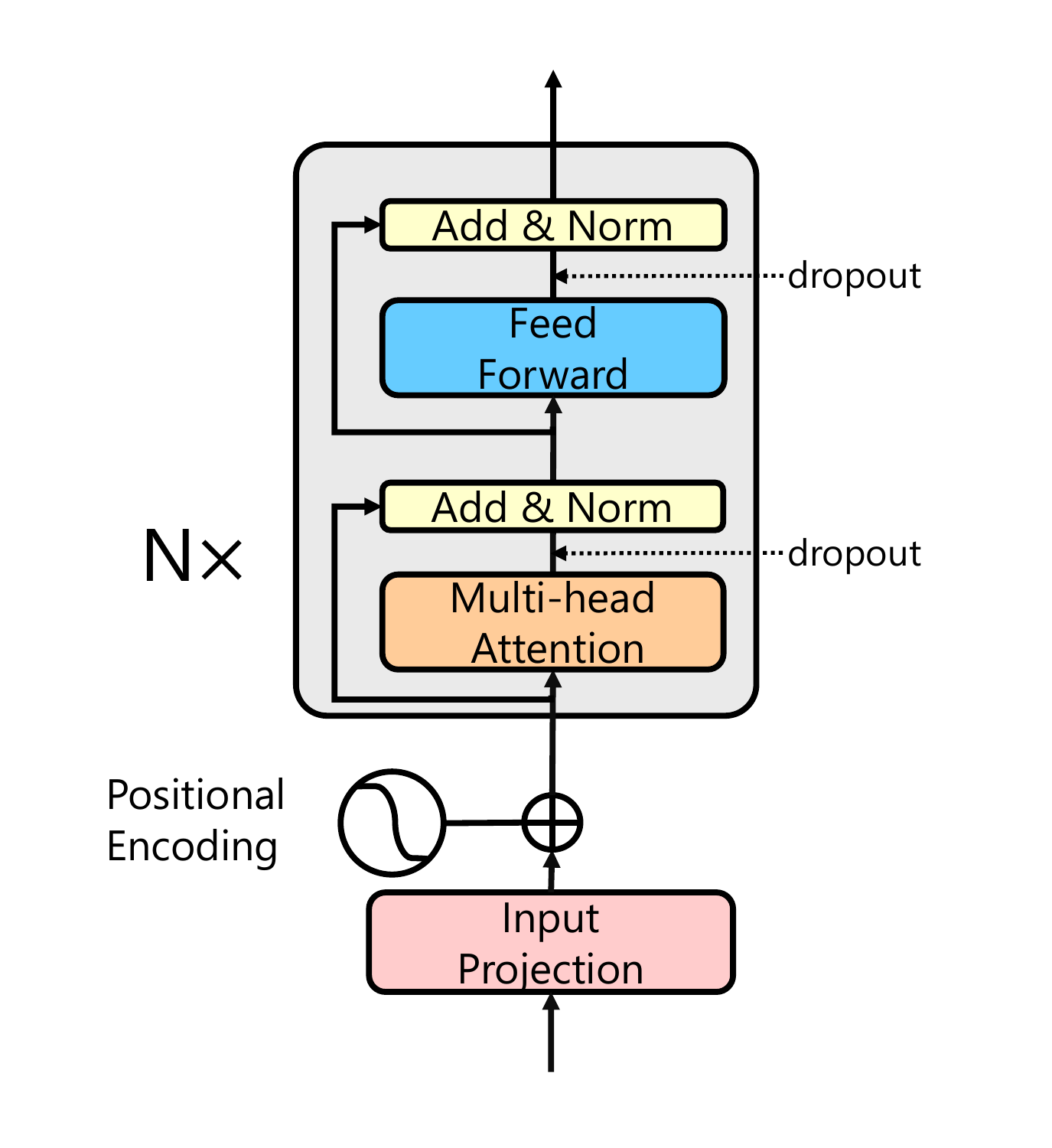}
         \caption{Transformer architecture.}
         \label{fig:transformer}
     \end{subfigure}
     \hfill
     \begin{subfigure}[b]{0.55\linewidth}
         \centering
         \includegraphics[width=1.0\textwidth]{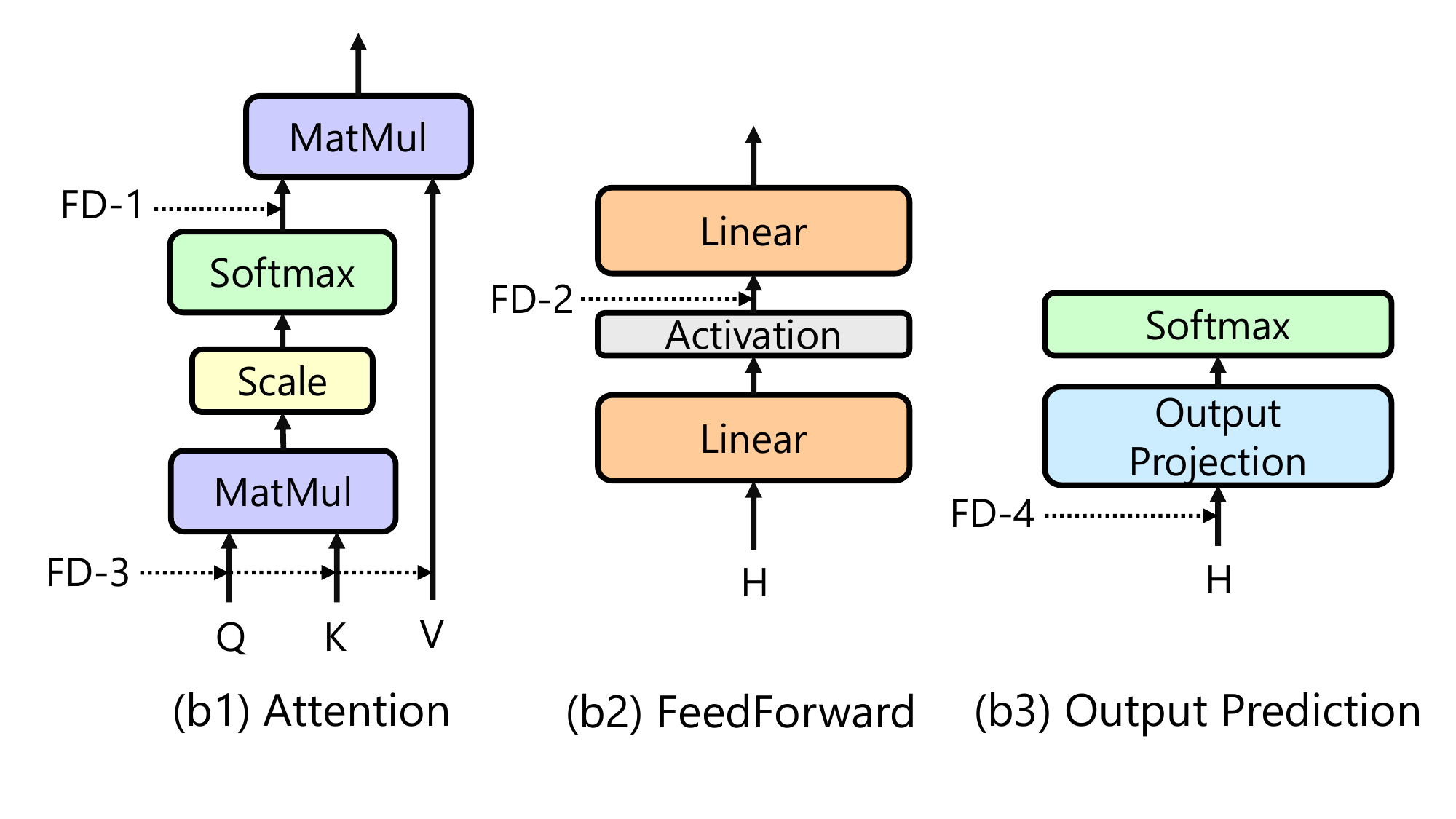}
         \caption{Structure and overview of feature dropout.}
         \label{fig:feature_dropout}
     \end{subfigure}
        \caption{Transformer structure and feature dropout applied in different Transformer components.}
        \label{fig:transformer_feature_dropout}
\end{figure*}

% \begin{figure}[!htbp]
% \centering
% \includegraphics[width=0.95\linewidth]{figs/transformer.png}
% \caption{Transformer architecture.}
% \label{fig_transformer}
% \end{figure}

We first theoretically analyze different regularization roles played by the three dropout techniques, and we show they can improve the generalization ability from different aspects. Then, we provide empirical evaluations of the $\mathtt{UniDrop}$ approach. We conduct experiments on neural machine translation with $8$ translation datasets, and text classification task with $8$ benchmark datasets. On both sequence generation and classification tasks, experimental results show that the three dropouts in $\mathtt{UniDrop}$ can jointly improve the performance of Transformer.

The contributions of this paper can be summarized as follows:
\begin{itemize}
    \item We introduce $\mathtt{UniDrop}$, which unites three different dropout techniques into a robust one for Transformer, to jointly improve the performance of Transformer without additional computational cost and prior knowledge.
    \item We theoretically demonstrate that the three dropouts, i.e., feature dropout, structure dropout, and data dropout play different roles in preventing Transformer from overfitting and improving the robustness of the model.
    \item Extensive results indicate that Transformer models with $\mathtt{UniDrop}$ can achieve strong or even SOTA performances on sequence generation and classification tasks. Specifically, around $1.5$ BLEU improvement on IWSLT14 translation tasks, and better accuracy for classification even using strong pre-trained model RoBERTa as backbone.
\end{itemize}

\section{Background}
\label{sec:background}
Feature dropout (FD) and structure dropout (SD) are highly coupled with model architecture. Therefore, we briefly recap Transformer and refer the readers to~\citet{vaswani2017attention} for details.

As shown in Figure~\ref{fig:transformer}, Transformer is stacked by several identical blocks, and each block contains two sub-layers, which are multi-head self-attention layer and position-wise fully connected feed-forward layer. Each sub-layer is followed by an $\mathtt{AddNorm}$ operation that is a residual connection $\mathtt{Add}$~\cite{he2016deep} and a layer normalization $\mathtt{LN}$~\cite{ba2016layer}.
% Transformer consists of one positional encoding module and multi-layers transformer blocks. Each block has two sub-layers, respectively a multi-head self-attention mechanism and a position-wise fully connected feed-foward network. Besides, each sub-layer is followed by a $\mathtt{AddNorm}$ operation that is a residual connection~\cite{} and a layer normalization~\cite{}.

\noindent \textbf{Multi-head Attention} sub-layer consists of multiple parallel attention heads,  and each head maps the query $\mathbf{Q}$ and a set of key-value pairs $\mathbf{K}, \mathbf{V}$ to an output through a scale dot-product attention:
\begin{equation}
\small
    \mathrm{Attn}(\mathbf{Q}, \mathbf{K}, \mathbf{V}) = \mathrm{softmax}(\frac{\mathbf{QK}^\top}{\sqrt{d_k}})\mathbf{V},
    \label{eqn:att_format}
\end{equation}
where $d_k$ is the dimension of query and key, and $\frac{1}{\sqrt{d_k}}$ is a scaling factor. The outputs of these heads are then concatenated and projected again to result in the final values. 
% where $d_k$ is the dimension of query and key. The scaling factor $\frac{1}{\sqrt{d_k}}$ is used to prevent from the dot products growing large in magnitude with the increase of dimension. The outputs of the multiple heads are then concatenated to jointly attend to information from different representation subspaces at different positions.

\noindent \textbf{Position-wise Feed-Forward} sub-layer applies two linear transformations with an inner ReLU~\cite{nair2010rectified} activation:
\begin{equation}
\small
\mathtt{FFN}(\mathbf{x})=\max(0, \mathbf{xW_1}+\mathbf{b_1})\mathbf{W_2}+\mathbf{b_2}, 
\end{equation}
where $\mathbf{W}$ and $\mathbf{b}$ are parameters.

The output of each sub-layer is then followed with $\mathtt{AddNorm}$: $\mathtt{AddNorm(x)} = \mathtt{LN}(\mathtt{Add(x)})$. 
% \quad The second sub-layer applies a fully connected feed-forward network to each position separately and identically,  $\mathtt{FFN}(\mathbf{x})=\mathbf{W_2}\mathtt{ReLU}(\mathbf{W_1 x}+\mathbf{b_1})+\mathbf{b_2}$, where $\mathbf{W}$ and $\mathbf{b}$ are the matrices and bias of parameters.

% Finally, the operations of each layer in the Transformer architecture can be formalized as:
% \begin{equation}
%     \mathtt{AddNorm}(\mathtt{FFN}(\mathtt{AddNorm(\mathtt{MultiHead}(\mathbf{Q}, \mathbf{K}, \mathbf{V}))})).
%     \label{layer_operation}
% \end{equation}

\section{UniDrop}
\label{sec:unidrop}
In this section, we first introduce the details of the three different levels of dropout techniques we study, feature dropout, structure dropout and data dropout. Then we provide the theoretical analysis of these dropout methods on the regularization perspectives. Finally, we present our proposed $\mathtt{UniDrop}$ approach for training Transformer. 
% As we mentioned, $\mathtt{UniDrop}$ integrates the three levels of dropout methods, feature dropout, structure dropout, into Transformer, in this section, we first introduce the details of each dropout, and then we theoretically analyze the different regularization perspectives of these dropout methods. 
% $\mathtt{UniDrop}$ integrates three levels of dropout methods, respectively, feature dropout, structure dropout, and data dropout into transformer models. In this section, we introduce them in details and give a theoretical analysis of their effects on prevent from overfitting.

% \begin{figure}[!htbp]
% \centering
% \includegraphics[width=1.05\linewidth]{figs/feature_dropout.pdf}
% \caption{Structure and overview of feature dropout.}
% \label{fig:feature_dropout}
% \end{figure}

\subsection{Feature Dropout}
\label{sec:feature_dropout}
The feature dropout (FD), as a well-known regularization method, is proposed by~\newcite{srivastava2014dropout}, which is to randomly suppress neurons of neural networks during training by setting them to 0 with a pre-defined probability $p$. 

In practice, dropout is applied to the output of each sub-layer by default. Besides, Transformer also contains two specific feature dropouts for multi-head attention and activation layer of feed-forward network. In this work, we also explore their effects on the performance of Transformer.

\begin{itemize}
    \item \textbf{FD-1} (attention dropout): according to Equation~(\ref{eqn:att_format}), we can obtain attention weight matrix $\mathbf{A}=\mathbf{QK^\top}$ towards value sequence $\mathbf{V}$. Our FD-1 is applied to the attention weight $\mathbf{A}$.
    \item \textbf{FD-2} (activation dropout): FD-2 is employed after the activation function between the two linear transformations of $\mathtt{FFN}$ sub-layer.
\end{itemize}

In addition to the above FDs for Transformer, we still find the risk of overfitting in pre-experiments. Therefore, we further introduce another two feature dropouts into the model architecture:

% Despite the effectiveness of the above FDs for Transformer, we still find the risk of overfitting in the pre-experiments. Therefore, we further add the two feature dropouts below into the Transformer architecture, as shown in Figure~\ref{fig:three sin x}.

\begin{itemize}
    \item \textbf{FD-3} (query, key, value dropout): FD-1 is used to improve generalization of multi-head attention. However, it is directly applied to the attention weights $\mathbf{A}$, where drop value $\mathbf{A}(i,j)$ means ignore the relation between token $i$ and token $j$, thus a larger FD-1 means a larger risk of losing some critical information from sequence positions. To alleviate this potential risk, we add dropout to query, key, and value before the calculation of attention.
    % there has been FD-1 used to improve generalization of multi-head attention in the Transformer models. Unfortunately, FD-1 is directly applied to the attention weights $\mathbf{\alpha}$, where each dimension corresponds to an objective position. Thus a larger attention dropout means a large risk of losing some key information from relevant positions. To address the issue, we add dropout to query, key, and value to further improve the capacity of multi-head attention mechanism.
    
    \item \textbf{FD-4} (output dropout): we also apply dropout to the output features before linear transformation for softmax classification. Specifically, when dealing with sequence-to-sequence tasks such as machine translation, we add FD-4 to the output features of the last layer in the Transformer decoder, otherwise the last layer of the Transformer encoder.
    % can add some notations such M to vector representation
\end{itemize}

The positions of each feature dropout applied in Transformer\footnote{We also explored other positions for feature dropout, but their performances are not so good (see Appendix \ref{sec:appendix_feature_dropout}).} are shown in Figure~\ref{fig:feature_dropout}. 

% \begin{algorithm}[t]
% 	\caption{Two-Stages Data Dropout}
% 	\label{alg:algorithm}
% 	\textbf{Input}: a sequence $s$; keep rate $p_k$ of the original sequence; data dropout rate $p$\\
% 	\textbf{Output}: new sequence $s'$ after applying data dropout to $s$
% 	\begin{algorithmic}[1] %[1] enables line numbers
% 	    \IF {the model is not in training stage}
% 		\STATE return $s$
% 		\ENDIF
% 	    \STATE Randomly sample $\xi_k: \xi_k \sim \mathrm{Bernoulli}(p_k)$
% 	    \IF {$\xi_k > 0.5$}
% 		\STATE Randomly sample $\xi: \xi \sim \mathrm{Bernoulli}(p)$
% 		\STATE Create a mask vector $\mathbf{m}$ of the sample length as the sentence $s$ according to $\xi$
% 		\STATE Apply the mask: $s' = s \times {\mathbf{m}}$
% 		\ELSE
% 		\STATE $s' = s$
% 		\ENDIF
% 		\STATE return $s'$
% 	\end{algorithmic}
% 	\label{opedecoding}
% \end{algorithm}

\subsection{Structure Dropout}
There are three structure dropouts, respectively $\mathtt{LayerDrop}$~\cite{fan2019reducing}, $\mathtt{DropHead}$~\cite{zhou2020scheduled} and $\mathtt{HeadMask}$~\cite{DBLP:journals/corr/abs-2009-09672}, which are specifically designed for Transformer.
% There are two existing structure dropout techniques specifically designed for the Transformer architecture, respectively $\mathtt{LayerDrop}$~\cite{fan2019reducing} and $\mathtt{DropHead}$~\cite{zhou2020scheduled}. 

 Some recent studies~\cite{DBLP:conf/acl/VoitaTMST19,DBLP:conf/nips/MichelLN19} show multi-head attention mechanism is dominated by a small portion of attention heads. To prevent domination and excessive co-adaptation between different attention heads,~\citet{zhou2020scheduled} and~\citet{DBLP:journals/corr/abs-2009-09672} respectively propose structured $\mathtt{DropHead}$ and $\mathtt{HeadMask}$ that drop certain entire heads during training. In contrast, $\mathtt{LayerDrop}$~\cite{fan2019reducing} is a higher-level and coarser-grained structure dropout. It drops some entire layers at training time and directly reduces the Transformer model size.

In this work, we adopt $\mathtt{LayerDrop}$ as the structure dropout to incorporate it into our $\mathtt{UniDrop}$. 
% We also tried $\mathtt{DropHead}$, while it fails to work in coordination with the feature dropout FD-1 and FD-3, thus we remove it from $\mathtt{UniDrop}$.

\subsection{Data Dropout}
Data dropout aims to randomly remove some words in the sentence with a pre-defined probability. It is often used as a data augmentation technique~\cite{DBLP:conf/emnlp/WeiZ19,xie2020unsupervised}. However, directly applying vanilla data dropout is hard to keep the original sequence for training, which leads to the risk of losing high-quality training samples. To address this issue, we propose a \emph{two-stage data dropout strategy}.
Specifically, given a sequence, with probability $p_k$ (a hyperparameter lies in $(0,1)$), we keep the original sequence and do not apply data dropout. If data dropout is applied, for each token, with another probability $p$ (another hyperparameter lies in $(0,1)$), we will drop the token. 
% Specifically, given a sequence, we first decide whether to apply data dropout according to a Bernoulli distribution with a parameter $p_k > 0$ that controls the keep probability of original sequence. If applied, each token of the sequence is dropped independently following another Bernoulli distribution associated with a parameter $p > 0$ that controls the drop rate of tokens.

% Nevertheless, directly applying data dropout may be unsuitable for some semantics-sensitive tasks, such as natural language inference and machine translation, because it is hard to keep the original sequence during training for a long sentence with direct data dropout.

% To alleviate the above problem, we propose a two-stages data dropout strategy. Specifically, given a sequence, we first decide whether to apply data dropout or not according to a Bernoulli distribution associated with a parameter $p_k > 0$ that controls the keep probability of original sequence. If applied, each token of the sequence is dropped independently following another Bernoulli distribution associated with a parameter $p > 0$ that controls the drop rate of tokens.

\subsection{Theoretical Analysis}

In this section, we provide theoretical analysis for feature dropout, structure dropout and data dropout, to show their different regularization effects. We first re-formulate the three dropout methods. For some probability $p$ and layer representation $h\in\mathbb{R}^d$ (i.e., $h$ is the vector of outputs of some layer), we randomly sample a scaling vector $\xi\in\mathbb{R}^d$ with  each independent coordinate as follows:
\begin{equation}
{\small \xi_i=\left\{
\begin{aligned}
-1 \quad&\textit{with probability p}  \\
\frac{p}{1-p} \quad&\textit{with probability 1-p}.
\end{aligned}
\right.}
\end{equation}
Here, $i$ indexes a coordinate of $\xi$, $i\in [1,...,d]$. Then feature dropout can be applied by computing {\small$$h_{fd}=(\boldsymbol{1}+\xi)\odot h,$$}where $\odot$ denotes element-wised product and $\boldsymbol{1}=(1,1,\cdots,1)'$. 

We use $F(h_{fd}(x))$ to denote the output of a model after dropping feature from a hidden layer and $\mathcal{L}$ to denote the loss function. Similar to \newcite{wei2020implicit}, we apply Taylor expansion to $\mathcal{L}$ and take expectation to $\xi$:%, we obtain
{\small\begin{align}
    &\mathbb{E}_{\xi}\mathcal{L}(F(h_{fd}(x)))=\mathbb{E}_{\xi}\mathcal{L}(F((\boldsymbol{1}+\xi)\odot h(x))) \notag\\
    &\approx\mathcal{L}(F(h(x))+\frac{1}{2}\mathbb{E}_{\xi}(\xi\odot h(x))^TD^2_h\mathcal{L}(x)(\xi\odot h(x)) \notag\\
    &=\mathcal{L}(F(h(x))+\frac{p}{2(1-p)}\sum_{j=1}^dD^2_{h_j,h_j}\mathcal{L}(x)\cdot h_j(x)^2,
\end{align}}where {$D_{h}^2\mathcal{L}$} is the Hessian matrix of loss with respect to hidden output {$h$} and {$D^2_{h_j,h_j}\mathcal{L}(x)$} is the $j$-th diagonal element of {$D_{h}^2\mathcal{L}$}. Expect the original loss $\mathcal{L}(F(h(x)))$, the above formula shows that feature dropout implicitly regularize the term $\sum_{j=1}^dD^2_{h_j,h_j}\mathcal{L}(x)\cdot h_j(x)^2$, which relates to the trace of the Hessian. 

For structure dropout, we use a 1-dim random scalar $\eta\in\mathbb{R}$ whose distribution is: $\eta=-1$ with probability $p$, and $\eta=0$ with probability $1-p$. The structure dropout is similarly applied by computing $h_{sd}=(1+\eta)\cdot h$. 

For input data $x\in \mathbb{R}^m$, here $x$ is a sequence of tokens and $m$ is the sequence length, we sample a random scaling vector $\beta\in \mathbb{R}^m$ with independent random coordinates where each coordinate is identically distributed as $\eta$. The input data after drop data becomes $x_{dd}=(\mathbf{1}+\beta)\odot x$.

Similar to feature dropout, we can obtain that data dropout implicitly optimizes the regularized loss as follows:
{\small$\mathcal{L}(F(h(x)))-p\cdot x^T\nabla_x\mathcal{L}(x)+ p\cdot\sum_{j=1}^mD^2_{x_j,x_j}\mathcal{L}(x)\cdot x_j^2$}, and structure dropout implicitly optimizes the regularized loss: {\small$\mathcal{L}(F(h(x)))-p\cdot h(x)^T\nabla_h\mathcal{L}(x)+p\cdot\sum_{i,j=1}^mD^2_{h_i,h_j}\mathcal{L}(x)\cdot h_i(x)h_j(x)$}, where $D^2_{h_i,h_j}\mathcal{L}(x)$ is the $(i,j)$-th element in Hessian matrix $D_{h}^2\mathcal{L}$.

\paragraph{Interpretation} From the above analysis, we can conclude that feature dropout, structure dropout and data dropout regularize different terms of the model, and they can not be replaced by each other. 
(1) Because the hidden output will be normalized by layer normalization, the term $h(x)^T\nabla_h\mathcal{L}(x)$ equals to zero according to Lemma 2.4 in \newcite{arora2018theoretical}. Therefore, structure dropout implicitly regularizes the term $\sum_{i,j=1}^mD^2_{h_i,h_j}\mathcal{L}(x)$. Hence, structure dropout can regularize the whole elements of Hessian of the model with respect to hidden output, while feature dropout only regularizes the diagonal elements of the Hessian. Thus, integrating structure dropout and feature dropout can regularize every component of Hessian with emphasizing the diagonal elements of the Hessian. (2) Since $x$ is also normalized, the term $x^T\nabla_x\mathcal{L}(x)$ equals to zero according to Lemma 2.4 in \newcite{arora2018theoretical}. Different from feature dropout and structure dropout, data dropout regularizes Hessian of loss with respect to input data.  

Regularizing Hessian matrix with respect to both input and hidden output can improve model robustness and hence the generalization ability. We put more details in Appendix \ref{sec:appendix_theoretic}.

\subsection{UniDrop Integration}

\begin{figure}[!tbp]
\centering
\includegraphics[width=0.85\linewidth]{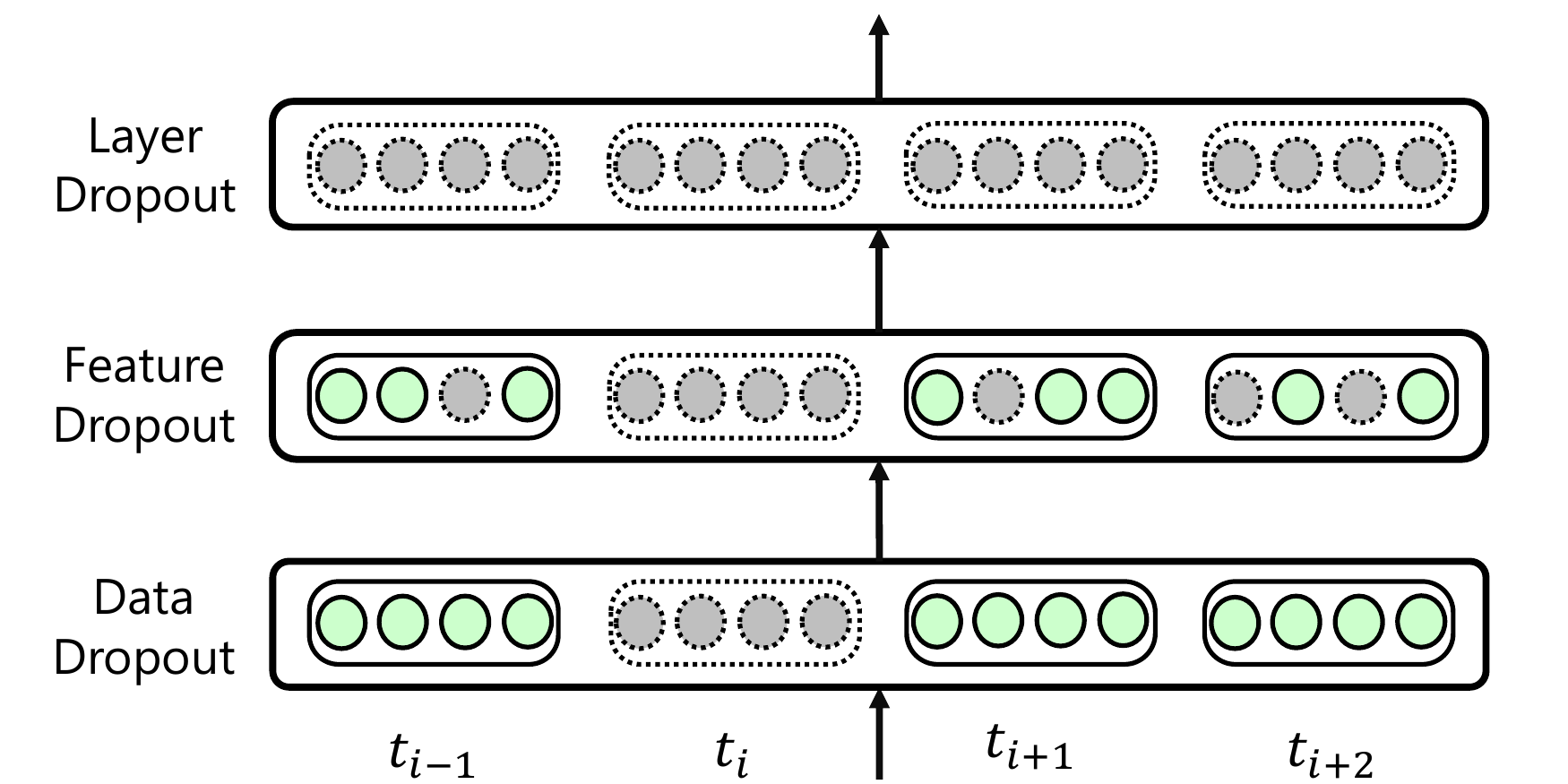}
\caption{Different dropout components in $\mathtt{UniDrop}$. The gray positions denote applying the corresponding dropout.}
\label{fig:unidrop}
\end{figure}

From the above theoretical analysis, the three dropout techniques are performed in different ways to regularize the training of Transformer, each with unique property to improve the model generalization. Therefore, we introduce $\mathtt{UniDrop}$ to take the most of each dropout into Transformer. The overview of $\mathtt{UniDrop}$ is presented in Figure \ref{fig:unidrop}. 

To better view each dropout in a model forward pass, we only show a three layers of architecture in Figure~\ref{fig:unidrop}, and each layer with one specific dropout technique. The data dropout is applied in the input layer by dropping out some word embeddings (e.g., embedding of word $t_i$ is dropped). In the middle layer, the feature dropout randomly drops several neurons in each word representations (e.g., the third neurons of word $t_{i-1}$ is dropped). The last layer is directly dropped out through layer dropout\footnote{Except the data dropout is only applied in the input layer, feature/structure dropout can be applied in each layer.}.

\section{Experiments}

\begin{table*}[!htbp]
    \centering
	\resizebox{1.0\textwidth}{!}
	{
		\centering
		\begin{tabular}{l | c c c c c c c c| c c}
			\hline
            & En$\to$De & De$\to$En & En$\to$Ro & Ro$\to$En & En$\to$Nl & Nl$\to$En & Nn$\to$Pt-br & Pt-br$\to$En & Avg. & $\triangle$ \\
			\hline
			Transformer & 28.67 & 34.84 & 24.74 & 32.14 & 29.64 & 33.28 & 39.08 & 43.63 & 33.25 & -\\
			\hline
			+FD & 29.61 & 36.08 & 25.45 & 33.12 & 30.37 & 34.50 & 40.10 & 44.74 & 34.24 & +0.99\\
			+SD & 29.03 & 35.09 & 25.03 & 32.69 & 29.97 & 33.94 & 39.78 & 44.02 & 33.69 & +0.44\\
			+DD & 28.83 & 35.26 & 24.98 & 32.76 & 29.72 & 34.00 & 39.50 & 43.71 & 33.59 & +0.34\\
			\hline
			+$\mathtt{UniDrop}$ & \textbf{29.99} & \textbf{36.88} & \textbf{25.77} & \textbf{33.49} & \textbf{31.01} & \textbf{34.80} & \textbf{40.62} & \textbf{45.62} & \textbf{34.77} & +1.52\\
			\hline
			w/o FD & 29.24 & 35.68 & 25.18 & 33.17 & 30.16& 33.90 & 39.97 & 44.81 & 34.01 & +0.76\\
			w/o SD & 29.92 & 36.70 & 25.59 & 33.26 & 30.55 & 34.75 & 40.45 & 45.60 & 34.60 & +1.35 \\
			w/o DD & 29.76 & 36.38 & 25.44 & 33.26 & 30.86 & 34.55 & 40.37 & 45.27 & 34.49 & +1.24\\
			\hline
	\end{tabular} }
	\caption{Machine translation results of the standard Transformer and our models on various IWSLT14 translation datasets. The ``+FD'', ``+SD'', ``+DD'', and ``+$\mathtt{UniDrop}$'' denotes applying the feature dropout, structure dropout, data dropout, or $\mathtt{UniDrop}$ to the standard Transformer. The ``w/o FD'', ``w/o SD'' and ``w/o DD'' respectively indicate the removal of the feature dropout, structure dropout, or data dropout from the model Transformer+$\mathtt{UniDrop}$. Avg. and $\triangle$ denote the average results of the 8 translation tasks and improvements compared with the standard Transformer. Best results are in bold.} 
	\label{mt_mainresults}
\end{table*}

We conduct experiments on both sequence generation and classification tasks, specifically, neural machine translation and text classification, to validate the effectiveness of $\mathtt{UniDrop}$ for Transformer.

\subsection{Neural Machine Translation}

In this section, we introduce the detailed settings for the neural machine translation tasks and report the experimental results.

\subsubsection{Datasets}
We adopt the widely acknowledged IWSLT14 datasets\footnote{\url{https://wit3.fbk.eu/mt.php?release=2014-01}} with multiple language pairs, including English$\leftrightarrow$German (En$\leftrightarrow$De), English$\leftrightarrow$Romanian (En$\leftrightarrow$Ro), English$\leftrightarrow$Dutch (En$\leftrightarrow$Nl), and English$\leftrightarrow$Portuguese-Brazil (En$\leftrightarrow$Pt-br), a total number of 8 translation tasks. Each dataset contains about 170k$\sim$190k translation data pairs. The datasets are processed by Moses toolkit\footnote{\url{https://github.com/moses-smt/mosesdecoder/tree/master/scripts}} and byte-pair-encoding (BPE)~\cite{DBLP:conf/acl/SennrichHB16a} is applied to obtain subword units. The detailed statistics of datasets are shown in Appendix \ref{sec:class_data_stat}.

\subsubsection{Model}
% Since all the language pairs contain about 100k$\sim$200k data pairs, 
We use the \texttt{transformer\_iwslt\_de\_en} configuration\footnote{\url{https://github.com/pytorch/fairseq}} for all Transformer models. Specifically, the encoder and decoder both consist of $6$ blocks. The source and target word embeddings are shared for each language pair. The dimensions of embedding and feed-forward sub-layer are respectively set to $512$ and $1024$, the number of attention heads is $4$. The default dropout (not our four feature dropout) rate is $0.3$  and weight decay is $0.0001$. All models are optimized with Adam~\cite{DBLP:journals/corr/KingmaB14} and the learning rate schedule is same as in~\citet{vaswani2017attention}. The weight of label smoothing~\cite{DBLP:conf/iclr/PereyraTCKH17} is set to $0.1$.

For the Transformer models with our $\mathtt{UniDrop}$, we set all feature dropout rates to $0.1$. The structure dropout $\mathtt{LayerDrop}$ is only applied to the decoder with rate $0.1$. For the data dropout, the sequence keep rate $p_k$ and token dropout rate $p$ are respectively $0.5$ and $0.2$. The other settings are the same as the configuration of the baseline Transformer.

To evaluate the model performance, we use beam search~\cite{sutskever2014sequence} algorithm to generate the translation results. The beam width is $5$ and the length penalty is $1.0$. The evaluation metric is the tokenized BLEU~\cite{DBLP:conf/acl/PapineniRWZ02} score with \texttt{multi-bleu.perl} script\footnote{\url{https://github.com/moses-smt/mosesdecoder/blob/master/scripts/generic/multi-bleu.perl}}. We repeat each experiment three times with different seeds and report the average BLEU.

\subsubsection{Results}

Table~\ref{mt_mainresults} shows the BLEU results of the Transformer baselines and models with different dropouts. Compared with baselines, we can see that the dropouts FD, SD, or DD all bring some improvements\footnote{The dropout rates of model Transformer+FD, Transformer+SD, Transformer+DD are tuned with IWSLT14 De$\to$En dev set and respectively set to $0.2$, $0.2$, $0.3$.}. This observation verifies the existence of overfitting in the Transformer. In contrast, our model Transformer+$\mathtt{UniDrop}$ achieves the most improvements across all translation tasks, which demonstrates the effectiveness of $\mathtt{UniDrop}$ for the Transformer architecture. To further explore the effects of the three different grained dropouts in $\mathtt{UniDrop}$, we conduct ablation studies and respectively remove the FD, SD, and DD from Transformer+$\mathtt{UniDrop}$. The results in Table~\ref{mt_mainresults} show that three ablated models obtain lower BLEU scores compared to the full model. This observation validates the necessity of them for $\mathtt{UniDrop}$. Among all ablation versions, the Transformer-$\mathtt{UniDrop}$ w/o FD obtains the least improvements. It is reasonable because FD actually contains four feature dropouts on different positions, which can effectively prevent Transformer from overfitting.

To show the superiority of $\mathtt{UniDrop}$, we also compare the Transformer+$\mathtt{UniDrop}$ with several existing works on the widely acknowledged benchmark IWSLT14 De$\to$En translation. These works improve machine translation from different aspects, such as the training algorithm design~\cite{DBLP:conf/icml/WangG019}, model architecture design~\cite{lu2019understanding,wu2018pay} and data augmentation~\cite{DBLP:conf/acl/GaoZWXQCZL19}. The detailed results are shown in Table~\ref{mt_comparsion}. We can see that the Transformer model with our $\mathtt{UniDrop}$ outperforms all previous works and achieve state-of-the-art performance, with $36.88$ BLEU score. Especially, it surpasses the BERT-fused NMT model~\cite{DBLP:conf/iclr/ZhuXWHQZLL20}, which incorporates the pre-trained language model BERT, by a non-trivial margin. We also show some comparisons on IWSLT14 En$\to$De, Ro$\to$En, and Nl$\to$En translations, the results are shown in Table~\ref{mt_comparsion_other}.

According to the above results, $\mathtt{UniDrop}$ successfully unites the FD, SD, and DD, and finally improves the performance of Transformer on neural machine translation tasks, without any additional computation costs and resource requirements.

\begin{table}[!tbp]
    \small
    \centering
	%\resizebox{0.9\textwidth}{!}
	{
		\centering
		\begin{tabular}{l c}
		    \hline
		    Approaches & BLEU \\
			\hline
			Adversarial MLE~\cite{DBLP:conf/icml/WangG019} & 35.18 \\
			DynamicConv~\cite{wu2018pay} & 35.20 \\
			Macaron~\cite{lu2019understanding} & 35.40 \\
			IOT~\cite{zhu2021iot} & 35.62 \\
% 			Joint Self-Attention~\cite{DBLP:journals/corr/abs-1905-06596} & 35.70 \\
			Soft Contextual Data Aug~\cite{DBLP:conf/acl/GaoZWXQCZL19} & 35.78 \\
			BERT-fused NMT~\cite{DBLP:conf/iclr/ZhuXWHQZLL20} & 36.11 \\
			MAT~\cite{DBLP:journals/corr/abs-2006-10270} & 36.22 \\
			MixReps+co-teaching~\cite{wu2020sequence} & 36.41 \\
			\hline
			Transformer & 34.84 \\
			+$\mathtt{UniDrop}$ & \textbf{36.88} \\
			\hline
           
	\end{tabular} }
	\caption{Comparison with existing works on IWSLT-2014 De$\to$En translation task.} 
	\label{mt_comparsion}
\end{table}

\begin{table}[!tbp]
    \small
    \centering
	\resizebox{1.0\linewidth}{!}
	{
		\centering
		\begin{tabular}{l| c c c}
		    \hline
		    Approaches & En$\to$De & Ro$\to$En & Nl$\to$En \\
			\hline
% 			Adversarial MLE~\cite{DBLP:conf/icml/WangG019} & 28.43 \\
% 			DynamicConv~\cite{wu2018pay} & 29.70 \\
% 			Macaron~\cite{lu2019understanding} &  \\
% 			Soft Contextual Data Aug~\cite{DBLP:conf/acl/GaoZWXQCZL19} &  \\
% 			BERT-fused NMT~\cite{DBLP:conf/iclr/ZhuXWHQZLL20} & 30.45 \\
			MAT~\cite{DBLP:journals/corr/abs-2006-10270} & 29.90 & - & -\\
			MixReps+co-teaching~\cite{wu2020sequence} & 29.93 & 33.12 & 34.45\\
			\hline
			Transformer & 28.67 & 32.14 & 33.38\\
			+$\mathtt{UniDrop}$ & \textbf{29.99} & \textbf{33.49} & \textbf{34.80} \\
			\hline
           
	\end{tabular} }
	\caption{Comparison with existing works on IWSLT-2014 En$\to$De, Ro$\to$En, and Nl$\to$En translation tasks.} 
	\label{mt_comparsion_other}
\end{table}

\subsection{Text Classification}

We also conduct experiments on text classification tasks to further demonstrate the effectiveness of $\mathtt{UniDrop}$ for the Transformer models.

\subsubsection{Datasets}

We evaluate different methods on the text classification task based on $8$ widely-studied datasets, which can be divided into two groups. The first group is from GLUE tasks~\cite{DBLP:conf/iclr/WangSMHLB19}, and they are usually used to evaluate the performance of the large-scale pre-trained language models after fine-tuning. The second group is some typical text classification datasets that are widely used in previous works~\cite{DBLP:conf/trec/VoorheesT99,DBLP:conf/acl/MaasDPHNP11,DBLP:conf/nips/ZhangZL15}. The statistics of all datasets are shown in Appendix \ref{sec:class_data_stat}. %IMDB dataset is binary film review classification task~\cite{DBLP:conf/acl/MaasDPHNP11}. Yelp and AG's News datasets are built by~\cite{DBLP:conf/nips/ZhangZL15}, respectively for sentiment classification and topic classification. TREC is a question classification dataset consisting of 6 question types~\cite{DBLP:conf/trec/VoorheesT99}. The statistics of all datasets are shown in Table~\ref{cls_statistics}.

\subsubsection{Model}
% To validate the effectiveness of UniDrop for Transformer, 
We employ $\text{RoBERTa}_\mathrm{{BASE}}$~\cite{liu2019roberta} as the strong baseline and fine-tune it on the text classification datasets. Different from $\text{BERT}_\mathrm{{BASE}}$~\cite{devlin2019bert}, $\text{RoBERTa}_\mathrm{{BASE}}$ is pre-trained with dynamic masking, full-sentences without NSP loss and a larger mini-batches. It has $12$ blocks, and the dimensions of embedding and $\mathtt{FFN}$ are $768$ and $3072$, the number of attention heads is $12$. When fine-tuning, we set the batch size to $32$ and the max epoch to $30$. Adam is applied to optimize the models with a learning rate of 1e-5 and a warm-up step ratio of $0.1$. We employ the polynomial decay strategy to adjust the learning rate. The default dropout and weight decay are both set to $0.1$.

When adding $\mathtt{UniDrop}$ to $\text{RoBERTa}_\mathrm{{BASE}}$, we empirically set feature dropout rate and $\mathtt{LayerDrop}$ rate to $0.1$. For data dropout, the sequence keep rate $p_k$ and token dropout rate $p$ are respectively $0.5$ and $0.1$. The other settings are the same as in the baseline $\text{RoBERTa}_\mathrm{{BASE}}$. We use the standard accuracy to evaluate different methods on text classification tasks.

\begin{table}[!tbp]
    \small
    \centering
	\resizebox{1.0\linewidth}{!}
	{
		\centering
		\begin{tabular}{l | c c c c}
			\hline
            & MNLI & QNLI & SST-2 & MRPC\\
			\hline
			BiLSTM+Attn, CoVe & 67.9 & 72.5 & 89.2 & 72.8\\
			BiLSTM+Attn, ELMo & 72.4 & 75.2 & 91.5 & 71.1\\
			$\text{BERT}_\mathrm{{BASE}}$ & 84.4 & 88.4 & 92.9 & 86.7  \\
			$\text{BERT}_\mathrm{{LARGE}}$ & 86.6 & 92.3 & 93.2 & 88.0  \\ 
			\hline
			$\text{RoBERTa}_\mathrm{{BASE}}$ & 87.1 & 92.7 & 94.7 & 89.0\\
			+$\mathtt{UniDrop}$ & \textbf{87.8} & \textbf{93.2} & \textbf{95.5} & \textbf{90.4}\\
			\hline
	\end{tabular} }
	\caption{Accuracy on GLUE tasks (dev set). The models BiLSTM+Attn, CoVe and BiLSTM+Attn, ELMo are from~\newcite{DBLP:conf/iclr/WangSMHLB19}. Best results are in bold.} 
	\label{cls_glue}
\end{table}

\begin{table}[!tbp]
    \small
    \centering
	\resizebox{1.0\linewidth}{!}
	{
		\centering
		\begin{tabular}{l | c c c c}
			\hline
            & IMDB & Yelp & AG & TREC\\
			\hline
			Char-level CNN & - & 62.05 & 90.49 & -\\
			VDCNN & - & 64.72 & 91.33 & - \\
			DPCNN & - & 69.42 & 93.13 & - \\
			ULMFiT & 95.40 & - & 94.99 & 96.40 \\
			$\text{BERT}_\mathrm{{BASE}}$ & 94.60 & 69.94 & 94.75 & 97.20  \\
			\hline
			$\text{RoBERTa}_\mathrm{{BASE}}$ & 95.7 & 70.9 & 95.1 & 97.6\\
			+$\mathtt{UniDrop}$ & \textbf{96.0} & \textbf{71.4} & \textbf{95.5} & \textbf{98.0}\\
			\hline
	\end{tabular} }
	\caption{Accuracy on the typical text classification datasets. Char-level CNN and VDCNN are from~\newcite{DBLP:conf/nips/ZhangZL15} and~\newcite{DBLP:conf/emnlp/ConneauKSBB17}, DPCNN and ULMFiT are from~\newcite{DBLP:conf/acl/JohnsonZ17} and~\newcite{DBLP:conf/acl/RuderH18}. Best results are in bold. } 
	\label{cls_zhang}
\end{table}

\subsubsection{Results}
Table~\ref{cls_glue} and Table~\ref{cls_zhang} respectively show the accuracy of different models on GLUE tasks and typical text classification datasets. 

Compared with the conventional BiLSTM and CNN based models, we can observe the pre-trained models, including ULMFiT, BERT, RoBERTa,  achieve obvious improvements on most datasets. 
% They show great superiority especially for tasks of language understanding, such as the inference tasks MNLI, QNLI and the paraphrase task MRPC. The improvements indicate the Transformer models pre-trained with large-scale corpora can capture substantial semantics and thus facilitate downstream text classification tasks. 
Benefiting from better training strategy, $\text{RoBERTa}_\mathrm{{BASE}}$ outperforms $\text{BERT}_\mathrm{{BASE}}$ and even $\text{BERT}_\mathrm{{LARGE}}$ on GLUE tasks.

We can see our proposed $\mathtt{UniDrop}$ further improve the performance $\text{RoBERTa}_\mathrm{{BASE}}$ on both small-scale and large-scale datasets. Specifically, $\mathtt{UniDrop}$ brings about $0.4$ improvements of accuracy on the typical text classification datasets from Table~\ref{cls_zhang}. In contrast, $\text{RoBERTa}_\mathrm{{BASE}}$+$\mathtt{UniDrop}$ achieves more improvements on GLUE tasks. The experimental results on the $8$ text classification benchmark datasets consistently demonstrate the facilitation of $\mathtt{UniDrop}$ for Transformer. We show more results and ablation study on text classification task in Appendix~\ref{sec:appendix_ablationstudy}.

\section{Analysis}

In this section, we use IWSLT14 De$\to$En translation as the analysis task to investigate the capability of $\mathtt{UniDrop}$ to avoid overfitting, as well as the effects of different dropout components and dropout rates on $\mathtt{UniDrop}$.

\subsection{Overfitting}
% \begin{figure}[!tbp]
% \centering
% \includegraphics[width=1.0\linewidth]{figs/loss.png}
% \caption{The training and development set losses of different models on IWSLT14 De$\to$En translation task.}
% \label{fig:loss}
% \end{figure}

\begin{figure}[!tbp]
\centering
\includegraphics[width=0.9\linewidth]{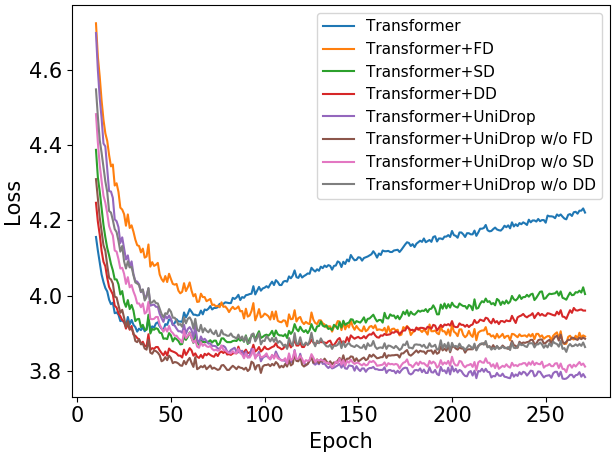}
\caption{The dev loss of different models on IWSLT14 De$\to$En translation task.}
\label{fig:loss}
\end{figure}
To show the superiority of $\mathtt{UniDrop}$ to prevent Transformer from overfitting, we compare the dev loss during training of Transformer, Transformer with each dropout technique, Transformer+$\mathtt{UniDrop}$, and ablated models of Transformer+$\mathtt{UniDrop}$. Figure~\ref{fig:loss} shows loss curves of different models.

We can observe that the standard Transformer is quickly overfitted during training, though it is equipped with a default dropout. In contrast, the feature dropout, structure dropout, and data dropout, as well as the combinations of any two dropouts (i.e., ablated models), greatly reduce the risk of overfitting to some extent. Among all compared models, our Transformer+$\mathtt{UniDrop}$ achieves the lowest dev loss and shows great advantage to prevent Transformer from overfitting. Besides, we also find that the dev loss of Transformer+$\mathtt{UniDrop}$ continuously falls until the end of the training. We stop it to keep training epochs of all models same for a fair comparison.

In Appendix \ref{sec:appendix_loss_curves}, we also plot the curves of training loss for the above models, together with the dev loss, to make a better understanding of the regularization effects from these dropout techniques.

\begin{table}[!tbp]
    \small
    \centering
	%\resizebox{0.9\textwidth}{!}
	{
		\centering
		\begin{tabular}{l| c c c}
		    \hline
		     & De$\to$En & En$\to$De & Ro$\to$En \\
			\hline
			Transformer & 34.84 & 28.67 & 32.14\\ 
			+$\mathtt{UniDrop}$ & 36.88 & 29.99 & 33.49\\
			\hline
			w/o FD-1 & 36.72 & 29.84 & 33.33\\
			w/o FD-2 & 36.57 & 29.76 & 33.28\\
			w/o FD-3 & 36.59 & 29.83 & 33.31\\
			w/o FD-4 & 36.65 & 29.59 & 33.24\\
			\hline
			w/o 2-stage DD & 36.61& 29.78 & 33.12\\
			\hline
           
	\end{tabular} }
    \caption{Ablation study of data dropout and different feature dropouts on IWSLT14 De$\to$En, En$\to$De, and Ro$\to$En translation tasks.} 
	\label{mt_ablation}
\end{table}

\begin{figure*}[!tbp]
\centering
\includegraphics[width=0.85\linewidth]{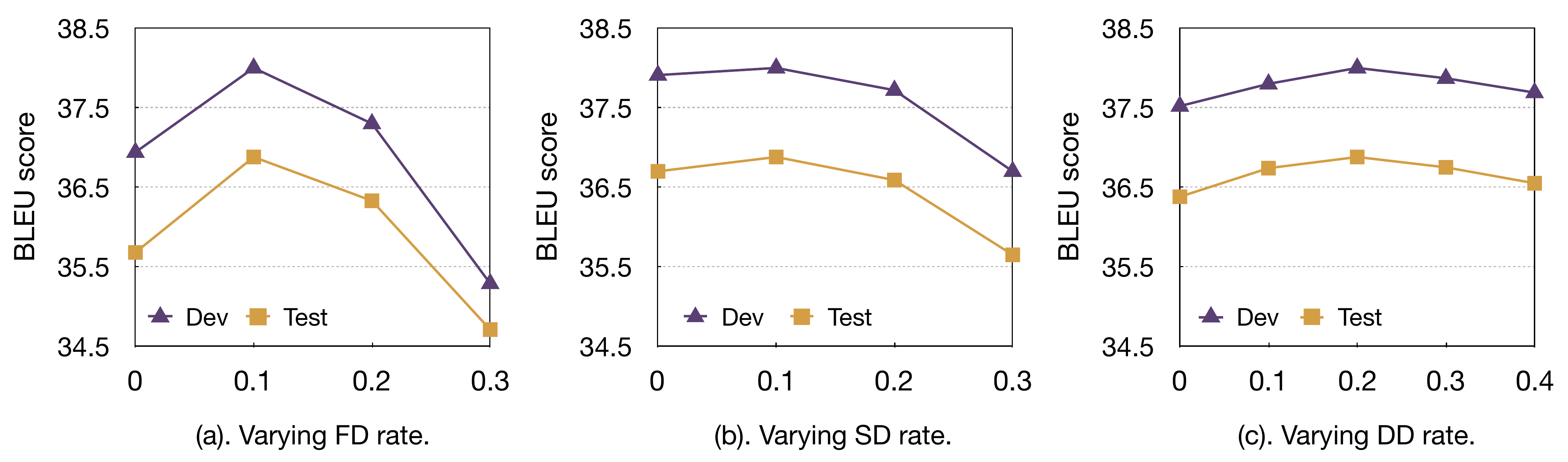}
\caption{The BLEU scores of Transformer+$\mathtt{UniDrop}$ on IWSLT14 De$\to$En translation dev set and test test, with varying the rates of FD, SD and DD respectively.}
\label{fig:varyingfd}
\end{figure*}

% \begin{figure*}[!tbp]
% \centering
% \includegraphics[width=0.9\textwidth]{figs/dropoutrates.pdf}
% \caption{The effects of respectively varying FD, SD and DD rates on Transformer+UniDrop with IWSLT-2014 De$\to$En translation task.}
% \label{fig:feature_dropout}
% \end{figure*}

\subsection{Ablation Study}
In Table~\ref{mt_mainresults}, we have presented some important ablation studies by removing FD, SD, or DD from $\mathtt{UniDrop}$. The consistent decline of BLEU scores demonstrates their effectiveness. Besides, we further investigate the effects of the two existing feature dropouts FD-1, FD-2, two new feature dropouts FD-3, FD-4, and our proposed two-stage data dropout strategy on Transformer models. The experimental results are shown in Table~\ref{mt_ablation}.

From Table~\ref{mt_ablation}, we can see the four ablation models removing FDs underperform the full model Transformer+$\mathtt{UniDrop}$, which means they can work together to prevent Transformer from overfitting. In multi-head attention module, FD-3 brings more BLUE improvement than FD-1. This comparison shows the insufficiency of only applying FD-1 for the Transformer architecture. The Transformer+$\mathtt{UniDrop}$ w/o 2-stage DD means we directly apply conventional data dropout to the sequence instead of our proposed 2-stage strategy. Compared with the full model, its performance also decreases. This shows the necessity of keeping the original sequence for data dropout.

\subsection{Effects of Different Dropout Rates}

To investigate the effects of FD, SD, and DD dropout rates on the $\mathtt{UniDrop}$, we respectively vary them based on the setting (FD=$0.1$, SD=$0.1$, DD=$0.2$). When varying one dropout component, we keep other dropout rates unchanged. Figure~\ref{fig:varyingfd} shows the corresponding results.

We can observe that the performance of each dropout for Transformer+$\mathtt{UniDrop}$ first increases then decreases when varying the dropout rates from small to large. Especially, varying the rate for FD dropout makes a more significant impact on the model performance since FD contains four feature dropout positions. In contrast, the DD is least sensitive to the dropout rate change, but it still plays a role in the model regularization.

\section{Related Work}

\subsection{Dropout}
Dropout is a popular regularization method for neural networks by randomly dropping some neurons during training~\cite{srivastava2014dropout}. Following the idea, there are abundant subsequent works designing specific dropout for specific architecture, such as StochasticDepth~\cite{DBLP:conf/eccv/HuangSLSW16}, DropPath~\cite{DBLP:conf/iclr/LarssonMS17}, DropBlock~\cite{DBLP:conf/nips/GhiasiLL18} for convolutional neural networks, Variational Dropout~\cite{DBLP:conf/nips/GalG16}, ZoneOut~\cite{DBLP:conf/iclr/KruegerMKPBKGBC17}, and Word Embedding Dropout~\cite{DBLP:conf/nips/GalG16} for recurrent neural networks. Recently, the Transformer architecture achieves great success in a variety of tasks. To improve generalization of Transformer, some recent works propose $\mathtt{LayerDrop}$~\cite{fan2019reducing}, $\mathtt{DropHead}$~\cite{zhou2020scheduled} and $\mathtt{HeadMask}$~\cite{DBLP:journals/corr/abs-2009-09672} as structured regularizations, and obtain better performance than standard Transformer. Instead of designing a specific dropout for Transformer, in this work, we focus on integrating the existing dropouts into one $\mathtt{UniDrop}$ to further improve generalization of Transformer without any additional cost.% We also theoretically demonstrate the three dropouts in \texttt{UniDrop} play different roles in regularizing Transformer.

\subsection{Data Augmentation}
Data augmentation aims at creating realistic-looking training data by applying a transformation to a sample, without changing its label~\cite{xie2020unsupervised}. In NLP tasks, data augmentation often refers to back-translation~\cite{sennrich2016improving}, word replacing/inserting/swapping/dropout~\cite{DBLP:conf/emnlp/WeiZ19,xie2020unsupervised}, etc. In this work, we adopt simple but effective word dropout as data level dropout in our $\mathtt{UniDrop}$. We, additionally, design a two-stage data dropout strategy.

\section{Conclusion}
In this paper, we present an integrated dropout approach, $\mathtt{UniDrop}$, to specifically regularize the Transformer architecture. The proposed $\mathtt{UniDrop}$ unites three different level dropout techniques from fine-grain to coarse-grain, feature dropout, structure dropout, and data dropout respectively. We provide a theoretical justification that the three dropouts play different roles in regularizing Transformer. Extensive results on neural machine translation and text classification datasets show that our Transformer+$\mathtt{UniDrop}$ outperforms the standard Transformer and various ablation versions. Further analysis also validates the effectiveness of different dropout components and our two-stage data dropout strategy. In conclusion, the $\mathtt{UniDrop}$ improves the performance and generalization of the Transformer without additional computational cost and resource requirement.

\section*{Acknowledgments}
The authors would like to thank the anonymous reviewers for their valuable comments. Xinyu Dai and Lijun Wu are the corresponding authors. This work was partially supported by the NSFC (No. 61976114,61936012) and National Key R\&D Program of China (No. 2018YFB1005102).

\bibliography{anthology,custom}
\bibliographystyle{acl_natbib}

%%%%%%%%%%%%%%%%%%%%%%%%%%%%
%%%%%%%%%%%%%%%%%%%%%%%%%%%%
%%%%%%%%%%%%%%%%%%%%%%%%%%%%

\clearpage
\appendix

\section{Appendix}
\subsection{Supplementary materials for theoretical analysis}
\label{sec:appendix_theoretic}
In this section, we explain why regularizing Hessian matrix with respect to input or hidden output can improve model robustness and generalization.

We use $D^{\alpha}_f\mathcal{L}$ to denote the $\alpha$-order derivatives of loss $\mathcal{L}$ with respect to $f$. If the hidden output is perturbed by $\epsilon$, i.e., $\tilde{h}=h+\epsilon$, the $k$-th output $F_k$ shifts to 
\begin{align}
F_k(h+\epsilon)=&F_k(h)+\epsilon^TJ_{F_k,h}\nonumber\\
&+\frac{1}{2}\epsilon^T(D_h^2 F_k(h))\epsilon+o(\epsilon^2),  \label{eq1} 
\end{align}where $J_{F_k,h}(x)$ is the Jacobian between hidden output $h$ and final output $F_k$. 

Structure dropout regularizes all elements in Hessian matrix $D_h^2\mathcal{L}$. For Hessian matrix of loss function, we have 
$D^2_h\mathcal{L}=J_{F,h}^T(D^2_F\mathcal{L})J_{F,h}+\sum_{k}(D_{F_k}\mathcal{L}) (D^2_hF_k(h))$. Thus, regularizing all elements in $D_h^2\mathcal{L}$ means regularizing both $J_{F,h}$ and $D_h^2F_k(h)$. As shown in Eq.\ref{eq1}, regularizing this two terms can make $|F_k(h+\epsilon)-F_k(h)|$ smaller. Therefore, the robustness of the model is improved and the generalization ability of the model can also be improved \cite{hoffman2019robust,jakubovitz2018improving}.

Feature dropout regularizes diagonal element of $D_h^2\mathcal{L}$. Using the approximation $D^2_h\mathcal{L}\approx J_{F,h}^T(D^2_F\mathcal{L})J_{F,h}$\cite{wei2020implicit}, regularizing diagonal elements $D_h^2\mathcal{L}$ equals to regularizing norm of Jacobian, i.e., $||J_{F,h}||_2$ if $D_F^2\mathcal{L}$ is roughly a diagonal matrix. For cross-entropy loss, $D_F^2\mathcal{L}=diag(z)-zz^T$, where $z$ is the probability vector predicted by the model encoding the distribution over output class labels, the matrix $D_F^2\mathcal{L}$ can be approximated by a diagonal matrix.  Thus, feature dropout mainly regularizes the first-order coefficient $J_{F_k,h}$ in Taylor expansion in Eq.\ref{eq1}, which is different from structure dropout. Since Jacobian is an essential quantity for the generalization \cite{wei2020implicit,hoffman2019robust}, emphasising this term is necessary for generalization although structure dropout can also regularize it.

Similar analysis can be applied to data dropout and we only need to replace hidden output $h$ to the input $x$.

\subsection{Statistics of Datasets}
\label{sec:class_data_stat}

\begin{table}[!tbp]
    %\small
    \centering
	%\resizebox{0.9\textwidth}{!}
	{
		\centering
		\begin{tabular}{c | c c c}
			\hline
            Datasets & Train & Dev & Test  \\
            \hline
            En$\leftrightarrow$De & 160k & 7k & 7k \\
            En$\leftrightarrow$Ro & 180k & 4.7k & 1.1k \\
            En$\leftrightarrow$Nl & 170k & 4.5k  & 1.1k \\
            En$\leftrightarrow$Pt-br & 175k  & 4.5k & 1.2k \\
			\hline
	\end{tabular} }
	\caption{Statistics for machine translation datasets.} 
	\label{mt_statistics}
\end{table}

\begin{table}[!tbp]
    % \small
    \centering
	%\resizebox{0.9\textwidth}{!}
	{
		\centering
		\begin{tabular}{l | c c c}
			\hline
            Datasets & Classes & Train & Dev\\
            \hline
            MNLI & 3 & 393k & 20k \\
            QNLI & 2 & 105k & 5.5k \\
            SST-2 & 2 & 67k & 0.9k \\
            MRPC & 2 & 3.7k & 0.4k \\
			\hline
		    Datasets & Classes & Train & Test\\
			\hline
			IMDB & 2 & 25k & 25k\\
			Yelp & 5 & 650k & 50k\\
			AG's News & 4 & 120k & 76k\\
			TREC & 6 & 5.4k & 0.5k\\
			\hline
	\end{tabular} }
	\caption{Statistics for text classification datasets.} 
	\label{cls_statistics}
\end{table}

Table~\ref{mt_statistics} and Table~\ref{cls_statistics} respectively show the statistics of machine translation and text classification benchmark datasets we used to evaluate the $\mathtt{UniDrop}$ for Transformer.

For machine translation tasks, the four language pairs all contain around 170k$\sim$190k training pairs. 
% Thus we use the same Transformer architecture \texttt{transformer\_iwslt\_de\_en} to configure all translation models. 

Text classification experiments are conducted in GLUE tasks~\cite{DBLP:conf/iclr/WangSMHLB19} and typical text classification benchmarks datasets~\cite{DBLP:conf/trec/VoorheesT99,DBLP:conf/acl/MaasDPHNP11,DBLP:conf/nips/ZhangZL15}. For GLUE tasks, we adopt the four datasets MNLI, QNLI, SST-2 and MRPC. They are used to evaluate the ability of models on language inference, sentiment classification and paraphrase detection. In typical text classification datasets, IMDB is binary film review classification task~\cite{DBLP:conf/acl/MaasDPHNP11}. Yelp and AG's News datasets are built by~\cite{DBLP:conf/nips/ZhangZL15}, respectively for sentiment classification and topic classification. TREC is a question classification dataset consisting of $6$ question types~\cite{DBLP:conf/trec/VoorheesT99}.

\subsection{Dropout Attempts}
\label{sec:appendix_feature_dropout}

\begin{table*}[!htbp]
    %\small
    \centering
	%\resizebox{0.9\textwidth}{!}
	{
		\centering
		\begin{tabular}{l |c}
		    \hline
		     & BLEU \\
			\hline
			Transformer & 34.84 \\ \hline
			+FD-1, FD-2 & 35.46 \\ \hline
			+FD-1, FD-2, FD-3 & 36.10 \\
			+FD-1, FD-2, QKV\_proj & 35.75 \\
			+FD-1, FD-2, FD-4 & 36.15 \\ 
			+FD-1, FD-2, LogitsDrop & 36.00 \\
			+FD-1, FD-2, FD-3, LogitsDrop & 36.06 \\ 
			+FD-1, FD-2, FD-3, FD-4 & 36.48 \\ 
			\hline
			+FD-1, FD-2, Encoder LayerDrop & 35.24 \\
			+FD-1, FD-2, Decoder LayerDrop & 35.99 \\
			+FD-1, FD-2, Encoder\&Decoder LayerDrop & 35.74 \\
			+FD-1, FD-2, EncoderDrop & 35.64 \\ 
			\hline
			+FD-1, FD-2, DD & 36.09 \\
			\hline
			+FD-1, FD-2, FD-3, FD-4, Decoder LayerDrop & 36.61 \\
			\hline
			+$\mathtt{UniDrop}$ & 36.88 \\
			\hline
           
	\end{tabular} }
	\caption{The results of different dropouts on IWSLT14 De$\to$En translation task.} 
	\label{mt_otherdropout}
\end{table*}

Besides the different dropout methods introduced in Section \ref{sec:unidrop}, we also tried some other dropouts. We first introduce their settings. The `QKV\_proj' applies dropout to query, key, and value after linear projection. In contrast, FD-3 is to add dropout to query, key, and value before projection. Similarly, `LogitsDrop' means that we use dropout after obtaining output logits from output projection layer. Compared to LogitsDrop, FD-4 directly applies dropout before the output projection layer. `EncoderDrop' means that we randomly drop the whole information of Transformer encoder with a probability and only use previous outputs to generate the next token during training. Obviously, it is a language modeling task when dropping the encoder. `Encoder LayerDrop' is that we apply $\mathtt{LayerDrop}$ only on the Transformer encoder.  Table~\ref{mt_otherdropout} shows the BLEU scores of different models on IWSLT-2014 De$\to$En translation task. All dropout rates are tuned within $[0.1, 0.2, 0.3, 0.4]$ according to the performance of the dev set.

FD-1 and FD-2 are two existing feature dropouts for Transformer. We first use them and achieve better BLUE scores than the standard Transformer, which demonstrates the existence of serious overfitting in Transformer model. On this basis, we try to add further feature dropout to prevent Transformer from overfitting. However, we can see that QKK\_proj achieves fewer improvements compared with FD-3. Similarly, LogitsDrop also underperforms FD-4. Therefore, we finally use FD-3 and FD-4 as our feature dropout components together with FD-1 and FD-2.

\begin{table}[!htbp]
    \small
    \centering
	\resizebox{1.0\linewidth}{!}
	{
		\centering
		\begin{tabular}{l | c c c c}
			\hline
            & MNLI & QNLI & SST-2 & MRPC\\
			\hline
			$\text{RoBERTa}_\mathrm{{BASE}}$ & 87.1 & 92.7 & 94.7 & 89.0\\
			+$\mathtt{UniDrop}$ & 87.8 & 93.2 & 95.5 & 90.4 \\
			%\hline
            w/o FD & 87.3 & 92.9 & 94.8 & 90.1 \\
            w/o SD & 87.5 & 93.1 & 95.1 & 89.5 \\
            w/o DD & 87.7 & 93.1 & 95.0 & 89.5 \\
			\hline
			$\text{RoBERTa}_\mathrm{{LAEGE}}$ & 89.8 & 94.3 & 96.3 & 90.4 \\
			+$\mathtt{UniDrop}$ & 90.2 & 94.8 & 96.6 & 91.4 \\
			%\hline
            w/o FD & 89.9 & 94.6 & 96.2 & 90.4 \\
            w/o SD & 90.0 & 94.6 & 96.3 & 90.7 \\
            w/o DD & 90.2 & 94.7 & 95.2 & 90.7 \\
			\hline
	\end{tabular} }
	\caption{Ablation Study on GLUE tasks (dev set). The ``w/o FD'', ``w/o SD'', ``w/o DD'' indicate respectively removing feature dropout, structure dropout, and data dropout from $\text{RoBERTa}_\mathrm{{BASE}}$+$\mathtt{UniDrop}$ or $\text{RoBERTa}_\mathrm{{LARGE}}$+$\mathtt{UniDrop}$.} 
	\label{cls_glue_ablation}
\end{table}

Among all structure dropout models, decoder LayerDrop outperforms all compared methods. In contrast, EncoderDrop only brings small improvements. Surprisingly, we can see that here the encoder LayerDrop actually has a negative effect on Transformer. 
% This may be attributed to the \texttt{transformer\_iwslt\_de\_en} consists of only 6 layers Transformer blocks and each layer encodes the important source information for translation. 
Thus we integrate the promising decoder LayerDrop as structured dropout component into $\mathtt{UniDrop}$. 
% Finally, $\mathtt{UniDrop}$ is further enhanced by data dropout and achieves the best performance on IWSLT14 De$\to$En translation task.

\subsection{Loss Curves}
\label{sec:appendix_loss_curves}

\begin{figure*}[!tbp]
\centering
\includegraphics[width=0.7\linewidth]{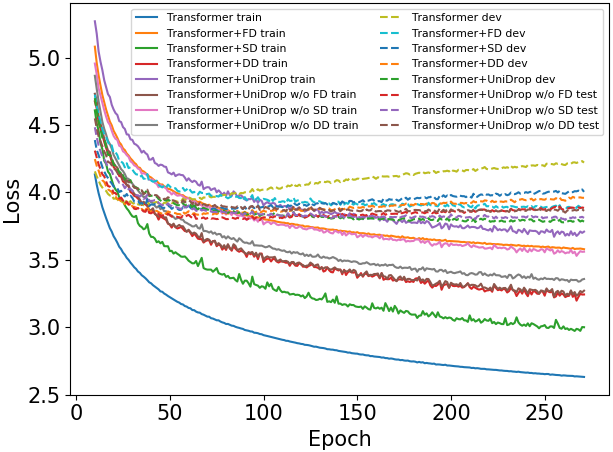}
\caption{The training and dev loss of different models on IWSLT14 De$\to$En translation task.}
\label{fig:trainingdevloss}
\end{figure*}

Figure~\ref{fig:trainingdevloss} shows the loss curves of different models during training. Overall, we can see that our Transfomer+$\mathtt{UniDrop}$ obtains the minimal gap of training loss and dev loss compared with other dropout models and the standard Transformer. This observation shows the better capability of $\mathtt{UniDrop}$ to prevent Transformer from overfitting. Benefitting from the advantage, Transfomer+$\mathtt{UniDrop}$ achieves the best generalization and dev loss on IWSLT14 De$\to$En translation task.

\subsection{Ablation Study on Text Classification}
\label{sec:appendix_ablationstudy}

Table~\ref{cls_glue_ablation} show the accuracy of standard $\text{RoBERTa}_\mathrm{{BASE}}$ and $\text{RoBERTa}_\mathrm{{LARGE}}$, the models with $\mathtt{UniDrop}$ and corresponding ablated models on GLUE tasks. Compared the base models $\text{RoBERTa}_\mathrm{{BASE}}$ and $\text{RoBERTa}_\mathrm{{LARGE}}$, we can observe that $\mathtt{UniDrop}$ further improves their performance on text classification tasks. After removing FD, SD, or DD from $\mathtt{UniDrop}$, the corresponding accuracy has decreased more or less. The consistent declines again demonstrate the necessity of the feature dropout, structure dropout and data dropout for $\mathtt{UniDrop}$.

\end{document}